\documentclass[10pt,a4paper,twocolumn]{article}

\usepackage[utf8]{inputenc}
\usepackage[T1]{fontenc}
\usepackage{newtxtext,newtxmath}
\DeclareUnicodeCharacter{221E}{\ensuremath{\infty}}
\DeclareUnicodeCharacter{2010}{-}
\usepackage[left=19mm,right=19mm,top=18mm,bottom=18mm,columnsep=7mm]{geometry}
\usepackage{microtype}
\usepackage{amsmath,bm}
\usepackage{graphicx}
\usepackage{booktabs}
\usepackage{enumitem}
\usepackage{algorithm}
\usepackage{algpseudocode}
\usepackage{subcaption}
\usepackage{cite}
\usepackage{fancyhdr}
\usepackage{titlesec}
\usepackage[hidelinks]{hyperref}

\hypersetup{
  pdftitle={A Fast Robot Localization Method Based on Monte Carlo Sampling},
  pdfauthor={He Zhaoqing, Hang Li, Xian Pu, Li Wang}
}

\setlist{nosep,leftmargin=2em}

\titleformat{\section}{\large\bfseries}{\thesection}{0.9em}{}
\titleformat{\subsection}{\normalsize\bfseries}{\thesubsection}{0.9em}{}
\titlespacing*{\section}{0pt}{1.25ex plus .4ex minus .2ex}{.75ex}
\titlespacing*{\subsection}{0pt}{1.1ex plus .3ex minus .2ex}{.55ex}

\fancypagestyle{main}{%
  \fancyhf{}
  \fancyhead[L]{\small Journal of *****, Vol. XX, No. XX}
  \fancyhead[R]{\small October, 2021}
  \fancyfoot[C]{\thepage}

}
\fancypagestyle{first}{%
  \fancyhf{}
  \fancyfoot[C]{\thepage}

}
\begin{document}

\twocolumn[
\begin{center}
  {\LARGE\bfseries A Fast Robot Localization Method Based on Monte Carlo\\[2pt]
  Sampling\par}
  \vspace{9pt}
  {\normalsize He Zhaoqing\textsuperscript{*},\quad Hang Li,\quad Xian Pu,\quad Li Wang\par}
  \vspace{5pt}
  {\small China Academy of Information and Communications Technology, Beijing, 100191, China\par}
  \vspace{4pt}
  {\small \textsuperscript{*}Corresponding author. E-mail: \href{mailto:zhaoqinghe0618@163.com}{zhaoqinghe0618@163.com}\par}
\end{center}

\vspace{9pt}
\begin{center}
  {\bfseries Abstract}
\end{center}
\vspace{-2pt}
\begin{minipage}{0.93\textwidth}
\small
\noindent\textbf{Abstract:} Localization is an essential capability of intelligent robots. However, in real-world constrained environments, robots often cannot rely on external localization systems. Therefore, this paper proposes an online particle-filter localization method based on Monte Carlo sampling. The method is applied to the localization of a mobile robot in a maze environment and requires only four range sensors mounted on the front, rear, left, and right sides of the vehicle, without the need for any heading information. Simulation experiments demonstrate that even in a maze where local regions are highly similar, the vehicle can still achieve relatively accurate localization, thereby providing a fast and effective approach for robot localization in complex environments.

\vspace{4pt}
\end{minipage}
\vspace{10pt}
]
\thispagestyle{first}

\section{Introduction}

Robots are the carriers of artificial intelligence and should possess three essential elements: perception, motion, and cognition~\cite{ref01_russell2010_artificialintelligenceamodernapproach,ref02_siciliano2016_springerhandbookofrobotics}. Nevertheless, current robots generally do not yet exhibit genuine intelligence and instead provide only basic programming and operational functions. Despite this limitation, their advantages in replacing humans for high-risk work have become increasingly evident~\cite{ref03_murphy2014_disasterrobotics,ref04_thrun2005_probabilisticroboticsintelligent}. In regions where the Global Positioning System (GPS) is denied, such as disaster zones and caves, performing fast localization and navigation using on-board sensors is an important problem for improving robot intelligence~\cite{ref05_cadena2016_pastpresentandfutureof,ref06_kohlbrecher2011_aflexibleandscalableslam}. Closely related is the path-planning task, for which metaheuristic optimizers such as heuristic hybrid particle swarm optimization have been developed to compute feasible quadcopter trajectories efficiently in complex environments~\cite{ref07_lou2024_hhpsoaheuristichybridparticle}; however, reliable path execution still presupposes accurate self-localization.

At present, common localization systems can be divided into external localization, such as GPS and motion capture, and on-board localization, such as visual navigation and radar navigation~\cite{ref08_durrantwhyte2006_simultaneouslocalizationandmappi,ref09_fuentespacheco2012_visualsimultaneouslocalizationan}. Owing to the high computational cost and power consumption of visual navigation~\cite{ref10_murartal2015_orbslamaversatileand,ref11_engel2014_lsdslamlargescaledirect}, it is not well suited for robot exploration in unknown environments. In radar navigation, filtering techniques are generally adopted to achieve localization, such as the Kalman filter, the bootstrap filter, the $H_\infty$ filter, and the particle filter~\cite{ref12_kalman1960_anewapproachtolinear,ref13_gordon1993_novelapproachtononlinearnon,ref14_simon2006_optimalstateestimation,ref15_arulampalam2002_atutorialonparticlefilters}. Each filtering technique has its own advantages and disadvantages. The Kalman filter is well suited for handling Gaussian distributions~\cite{ref12_kalman1960_anewapproachtolinear,ref16_welch1995_introductionkalmanfilter}; the particle filter is chosen for large-scale environments~\cite{ref17_montemerlo2002_fastslamafactoredsolutionto}, because compared with the Kalman filter that has a complexity of $O(n^2)$, its computational complexity is logarithmic; the $H_\infty$ filter is used to improve the convergence of Simultaneous Localization and Mapping (SLAM) systems~\cite{ref18_ahn2012_robusthfilteringforslam}.

To achieve fast localization in unknown and complex environments, this paper selects the particle filter because it imposes relatively minimal requirements on the system (it requires neither that the system model equations be linear nor that the system noise be Gaussian)~\cite{ref19_doucet2009_particlefilteringtutorial}, and thus offers better algorithmic scalability and generality. In this work, the robot is placed in a complex maze, and its position and heading are inferred by the particle filter. Four sensors are mounted on the front, rear, left, and right sides of the robot, and these sensors are used to measure the perpendicular distance between the robot and the nearest wall. By comparing the observation data with the sampled data, the distribution of the pose estimate is revised, followed by state transition and resampling. Experiments show that an accurate localization of the robot can be obtained after several rounds of resampling.

Recently, the integration of large language models with multi-agent systems and reinforcement learning has opened new directions for autonomous agents and swarm intelligence~\cite{ref20_wang2024_asurveyonlargelanguage,ref21_zhao2023_asurveyoflargelanguage}. For example, language-model-based agents have been used to automate the configuration and training of multi-agent reinforcement learning pipelines for drone swarms~\cite{ref22_lou2024_agentstrainerautomaticallytraini}. These advances complement classical probabilistic localization by enriching high-level reasoning and coordination, while methods such as the one proposed here remain crucial for the low-level state estimation upon which higher-level autonomy depends.

The remainder of this paper is organized as follows. Section~2 introduces the theoretical foundation and operating mechanism of the particle filter; Section~3 evaluates the proposed algorithm in a simulation environment; and Section~4 draws the conclusion.

\section{Particle Filter}

The particle filter performs Bayesian filtering through non-parametric Monte Carlo sampling~\cite{ref23_doucet2001_sequentialmontecarlomethodsinpractice,ref24_liu1998_sequentialmontecarlomethodsfor}. Its core idea is to represent a distribution using random state particles sampled from the posterior probability. That is, it approximates the probability density function by finding a set of random samples propagating in the state space, and replaces the integration operation with the sample mean, thereby obtaining a process with the minimum-variance distribution of the state~\cite{ref25_li2014_fightsampledegeneracyandimpoveri}. Because the particle filter can effectively solve nonlinear dynamic system problems, it is widely applied in the fields of target tracking and localization~\cite{ref26_gustafsson2002_particlefiltersforpositioningnav,ref27_okuma2004_aboostedparticlefiltermultitarge}.

\subsection{Bayesian Filtering}

The essence of the localization problem is recursive state estimation, i.e., recursively estimating the states $x_1,x_2,\ldots,x_k$ from the sequential observations $y_1,y_2,\ldots,y_k$ at discrete time $k$, where the observation $y_k$ is a function of the state $x_k$, and the evolution of the state $x_k$ over time is generally described as a discrete or continuous Markov process~\cite{ref28_cappe2007_anoverviewofexistingmethods,ref29_doucet2000_onsequentialmontecarlosampling}:
\begin{equation}
x_k=f_k(x_{k-1},w_k),
\end{equation}
where $k$ is the discrete time and $w_k\in\mathbb{R}^{d_w}$ is the process noise.

The system observation equation is:
\begin{equation}
y_k=h_k(x_k,v_k),
\end{equation}
where $y_k$ is the observation value and $v_k$ is the observation noise.

The state equation and the observation equation together constitute the state-space model that describes the entire localization problem. In this model, the filter can obtain the filtering distribution $p(x_k\mid y_{1:k})$, whose solution can be described as a recursive naive Bayesian filter based on the law of total probability, consisting of two parts: prediction (3) and update (4)~\cite{ref30_chen2003_bayesianfilteringfromkalmanfilte}.
\begin{align}
p(x_k\mid y_{1:k-1})={}&\int p(x_k\mid x_{k-1}) \notag\\
&\quad p(x_{k-1}\mid y_{1:k-1})\,\mathrm{d}x_{k-1},
\end{align}
\begin{equation}
p(x_k\mid y_{1:k})=
\frac{p(y_k\mid x_k)p(x_k\mid y_{1:k-1})}
{\int p(y_k\mid x_k)p(x_k\mid y_{1:k-1})\,\mathrm{d}x_k}.
\end{equation}
where the state-transition probability distribution $p(x_k\mid x_{k-1})$ and the observation-likelihood function distribution $p(y_k\mid x_k)$ are obtained from the state equation (1) and the observation equation (2), respectively. Then, given the posterior filtering distribution $p(x_{k-1}\mid y_{1:k-1})$ at time $k-1$, the posterior probability distribution at time $k$ can be obtained as:
\begin{equation}
p(x_k\mid y_{1:k})=\int p(x_{k-1}\mid y_{1:k-1})q(x_k)\,\mathrm{d}x_{k-1},
\end{equation}
\begin{equation}
q(x_k)=\frac{p(y_k\mid x_k)p(x_k\mid x_{k-1})}
{\int p(y_k\mid x_k)p(x_k\mid y_{1:k-1})\,\mathrm{d}x_k}.
\end{equation}

In engineering practice, Eqs.~(4)--(5) have analytical solutions only under Gaussian distributions. For the robot localization problem, which belongs to nonlinear systems, an approximate computation is required~\cite{ref31_julier2004_unscentedfilteringandnonlineares,ref32_wan2000_unscentedkalmanfilter}. The particle filter used in this paper is a non-parametric approximation that represents the probability distribution with weighted discrete particles, also known as sequential Monte Carlo. It requires neither linearity of the state and observation equations nor that the noise be Gaussian or even independent~\cite{ref33_pitt1999_filteringviasimulationauxiliaryp,ref34_kitagawa1996_montecarlofilterandsmoother}.

\subsection{Monte Carlo Particle Filter}

For an arbitrary probability distribution $p(x_k)$, a Monte Carlo approximation can be performed using discrete particles:
\begin{equation}
p(x_k\mid y_{1:k})\approx\sum_{i=1}^{N_k}w_k^{(i)}\delta\!\left(x_k-x_k^{(i)}\right),
\end{equation}
where $x_k^{(i)}$ denotes the state of a particle, $w_k^{(i)}$ is the particle weight, $N_k$ is the total number of particles, and $\delta(\cdot)$ is the Dirac delta function~\cite{ref35_wilkin1981_principlesofquantummechanics}.

We perform sampling and resampling on this approximation, namely:

\noindent\textbf{Step 1: Sampling.} Based on the Bayesian posterior estimate at $k-1$ and the state-transition equation, the particle states are updated to obtain the prior distribution $p(x_k\mid y_{1:k-1})$. Generally, a certain distribution function must be designed as the sampling (proposal) function, which is then optimized through Step 2 to approximate the posterior probability distribution~\cite{ref29_doucet2000_onsequentialmontecarlosampling}.

\noindent\textbf{Step 2: Update.} Based on the latest observation information $y_k$, the particle weights are updated through the likelihood function, as shown in Eq.~(8).
\begin{equation}
w_k^{(i)}\propto w_{k-1}^{(i)}
\frac{p\!\left(y_k\mid x_k^{(i)}\right)p\!\left(x_k^{(i)}\mid x_{k-1}^{(i)}\right)}
{q\!\left(x_k^{(i)}\right)}.
\end{equation}
After all weights are updated, normalization is required:
\begin{equation}
\bar{w}_k^{(i)}=\frac{w_k^{(i)}}{\sum_{i=1}^{N_k}w_k^{(i)}}.
\end{equation}

\noindent\textbf{Step 3: Resampling.} Based on the principle of identical distribution, the particle set after weight updating is resampled to obtain a new particle set in which most particles have comparable (or even equal) weights, thereby overcoming the particle weight degeneracy problem~\cite{ref36_li2015_resamplingmethodsforparticlefilt,ref37_douc2005_comparisonofresamplingschemesfor}.

\section{Algorithm Test Results}

\subsection{Test Environment}

\begin{figure}[t]
  \centering
  \includegraphics[width=.72\columnwidth]{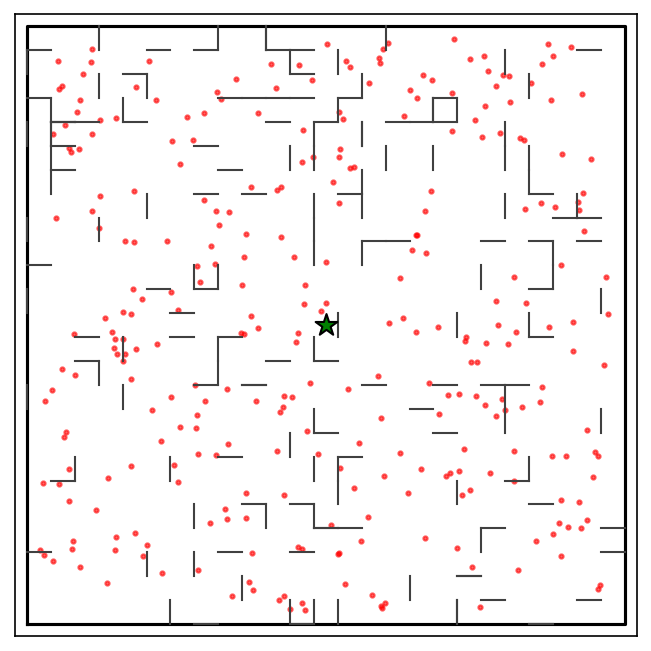}
  \caption{Scene map.}
  \label{fig:scene}
\end{figure}

In this paper, the robot is placed in a maze, as shown in Fig.~\ref{fig:scene}. Because the small turtle has strong demonstrative significance in the ROS (Robot Operating System) environment~\cite{ref38_quigley2009_rosanopensourcerobotoperating}, we choose the green turtle model to represent the robot. The robot can perceive the surrounding environment through four distance sensors mounted on the front, rear, left, and right sides. However, owing to its limited perception capability, it can only sense obstacles within a range of 5 meters, while the entire environment is 25 meters in both length and width. The goal of the algorithm is to quickly determine the robot's location on the map within a short time. In addition, the robot is equipped with an odometer that can measure the current local velocity and acceleration, and the red dots represent the sampled particles. Simulation platforms that mirror real-world dynamics, such as visual-reality many-agent environments for large-scale aerial systems~\cite{ref39_lou2023_airmavisualreality}, share a similar motivation of bridging simulated and physical deployment, although our focus here is on single-robot probabilistic localization.

\subsection{Algorithm Setup}

As shown in Algorithm~\ref{alg:mcpf}, the Monte Carlo Particle Filter (MCPF) proposed in this paper mainly consists of five steps:
\begin{enumerate}
  \item Randomly sample $N$ particles. In this paper, a uniform distribution is selected as the initial distribution, and each sampled particle contains both position and heading information.
  \item Read the particle's position on the map and the sensor measurement data, and update the particle weights according to the measurement equation.
  \item Normalize all computed weights.
  \item Resample the particles according to the normalized weights and update the particle distribution.
  \item Predict the state distribution of all particles based on the robot's previous state according to the state equation.
\end{enumerate}

\begin{algorithm}[t]
\caption{Monte Carlo Particle Filter}
\label{alg:mcpf}
\footnotesize
\begin{algorithmic}[1]
\Require Map data, robot sensor readings
\Ensure Current robot localization
\State Initialize all particles and their weights:
\State $X_0\sim\mathrm{Uniform};\ \hat{X}_0=\varnothing;\ w_0=1/N;\ j\gets0$
\While{not at end of the task}
  \State $Z_j\gets\mathrm{CurrentSensorData}$
  \State $W_j\gets\mathrm{SensorUpdate}(X_j,Z_j)$
  \State $\mathrm{Normalization}(W_j)$
  \State $\hat{X}_j=\langle X_j,W_j\rangle,\ X_{j+1}=\varnothing$
  \For{$n=0$ to $N-1$}
    \State draw $x_j^{[n]}$ from $\hat{X}_j$ with probability $W_j$
    \State $X_{j+1}\gets X_{j+1}+\mathrm{MotionUpdate}(x_j^{[n]},u_j)$
  \EndFor
  \State $W_j\gets1/N;\ j{+}{+}$
\EndWhile
\end{algorithmic}
\end{algorithm}

\subsection{Test Results and Analysis}

\begin{figure}[t]
  \centering
  \begin{subfigure}[b]{.47\columnwidth}
    \centering
    \includegraphics[width=\linewidth]{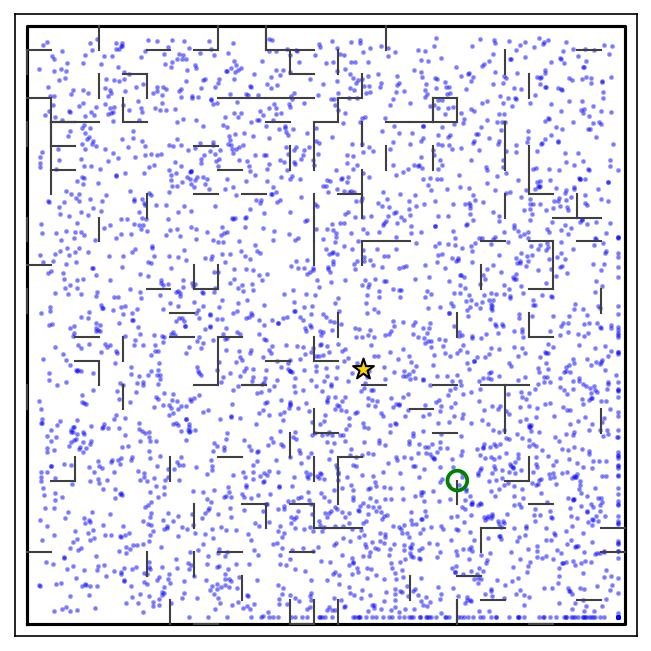}
    \caption{}
  \end{subfigure}\hfill
  \begin{subfigure}[b]{.47\columnwidth}
    \centering
    \includegraphics[width=\linewidth]{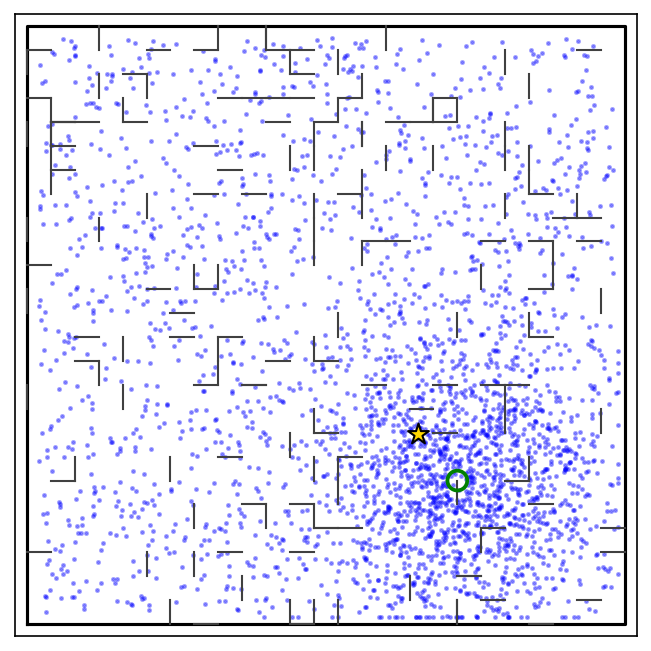}
    \caption{}
  \end{subfigure}
  \vspace{2pt}
  \begin{subfigure}[b]{.47\columnwidth}
    \centering
    \includegraphics[width=\linewidth]{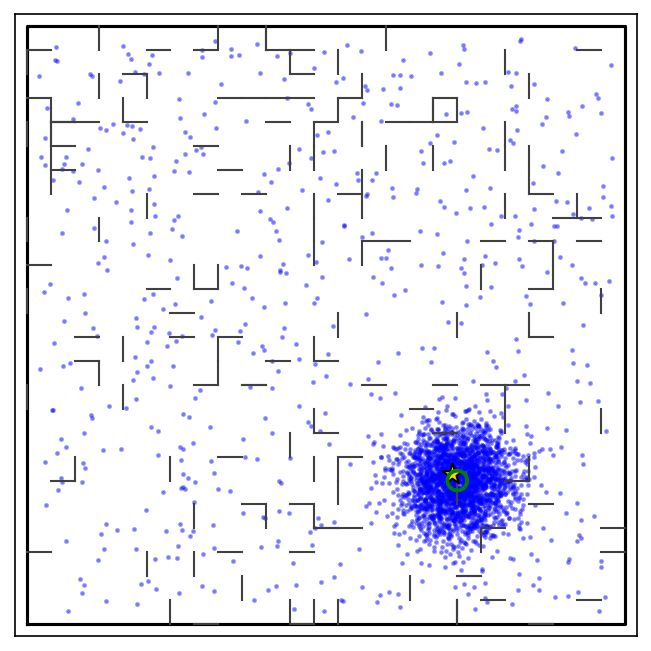}
    \caption{}
  \end{subfigure}\hfill
  \begin{subfigure}[b]{.47\columnwidth}
    \centering
    \includegraphics[width=\linewidth]{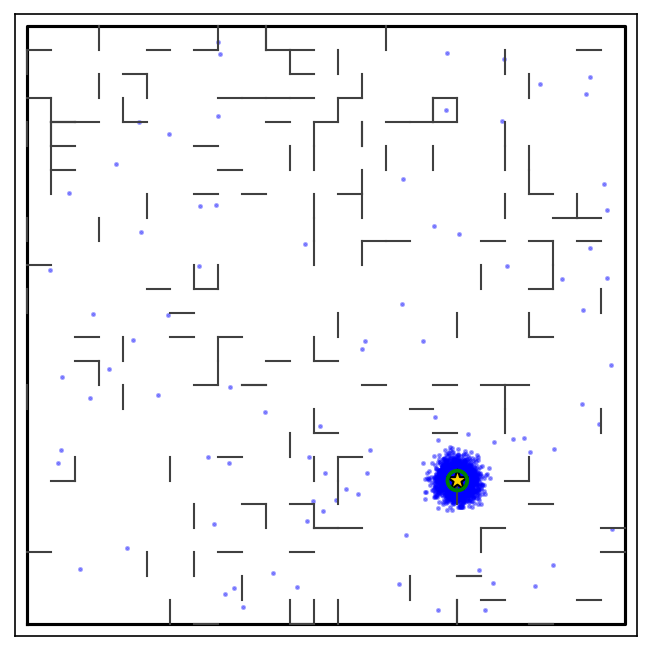}
    \caption{}
  \end{subfigure}
  \caption{Test results.}
  \label{fig:results}
\end{figure}

We implemented Algorithm~\ref{alg:mcpf} in Python and tested it in the environment described in Section~3.1; the test results are shown in Fig.~\ref{fig:results}. Initially, we scattered 3000 particles over the entire map using a uniform distribution, and the arrow of each particle indicates its heading. After a simple weighted summation over all particles, we marked the current predicted position of the particle filter, as shown by the yellow turtle. Because of the uniform distribution, the predicted result is roughly located at the center of the map.

Subsequently, we compute the observation equation for all sampled particles, update the weight of each particle, and after normalization form a new probability distribution, based on which we perform resampling. We then let the turtle move one step, and all sampled particles change their current pose according to the state equation; the result after movement is shown in Fig.~\ref{fig:results}(a). After the first iteration, although the sampling results are still scattered across various corners of the map, the predicted result is closer to the ground truth. We repeat the above steps: after the second iteration, the particles are no longer disordered but gradually converge to a few points, as shown in Fig.~\ref{fig:results}(b); after the third iteration, the predicted result is already very close to the ground truth, but a portion of the particles has not yet converged, as shown in Fig.~\ref{fig:results}(c); after the fourth iteration, the particles converge, and the predicted result is essentially consistent with the ground truth, as shown in Fig.~\ref{fig:results}(d).

\section{Conclusion}

As an important nonlinear Bayesian filtering method and statistical signal-processing tool, the particle filter is guided by probability and statistics theory, imposes loose requirements on the system model, and is particularly suitable for strongly nonlinear and non-Gaussian systems~\cite{ref15_arulampalam2002_atutorialonparticlefilters,ref40_gustafsson2010_particlefiltertheoryandpractice}. By combining Bayesian estimation and the Monte Carlo method, this paper successfully uses a particle filter to perform fast and accurate localization of a robot in a maze based on simple sensor readings. The test results show that the algorithm converges quickly and localizes accurately, and can be applied to real-world environments. Future work may explore the combination of such probabilistic estimators with learning-based and language-model-driven multi-agent frameworks, for instance task-activated knowledge-extension reasoning systems that let large language models plan and refine swarm behaviors~\cite{ref41_lou2024_talkerataskactivatedlanguage,ref42_zhang2023_buildingcooperativeembodiedagent}, in order to enable scalable autonomy in large-scale unmanned systems.

\clearpage

\nocite{ref43_oroojlooy2022_areviewofcooperativemulti,ref44_canese2021_multiagentreinforcementlearninga,ref45_ahn2022_doasicannot,ref46_bailey2006_slam_partii,ref47_gronauer2022_multiagentdeepreinforcementlearning,ref48_azar2021_dronedeepreinforcementlearninga,ref49_kober2013_reinforcementlearninginroboticsa,ref50_dellaert1999_montecarlolocalization,ref51_chen2023_deepreinforcementlearningforauto,ref52_li2023_deepreinforcementlearningforrobo,ref53_thrun2001_robustmontecarlolocalizationfor,ref54_wang2024_asurveyonrobotlocalization,ref55_zhu2021_deepreinforcementlearningbasedmo,ref56_tan2022_asurveyofmultirobotcooperative,ref57_wan2018_robotassistedpedestrianregulatio,ref58_nguyen2020_deepreinforcementlearningmultiagent,ref59_huang2022_languagemodelsaszeroshot,ref60_xiao2022_motionplanningandcontrolfor,ref61_singh2021_reinforcementlearninginroboticap,ref62_brown2020_languagemodelsarefewshot,ref63_kaufmann2023_championleveldroneracingusing,ref64_orr2023_multiagentdeepreinforcementlearn,ref65_fox2003_kldsampling,ref66_xu2023_creativerobottoolusewith,ref67_kuutti2021_deeplearningautonomousvehiclecontrol,ref68_grisetti2007_gridmappingraoblackwellized}

\bibliographystyle{IEEEtran}
\bibliography{references}

\end{document}